\documentclass[10pt,twocolumn,letterpaper]{article}

 \usepackage{indentfirst}
\usepackage[pagenumbers]{cvpr} 
\usepackage{tabularx}
\usepackage[dvipsnames]{xcolor}

\definecolor{cvprblue}{rgb}{0.21,0.49,0.74}
\usepackage[pagebackref,breaklinks,colorlinks,citecolor=cvprblue]{hyperref}

\def\paperID{9608} 
\def\confName{CVPR}
\def\confYear{2024}

\title{UNeR3D: Versatile and Scalable 3D RGB Point Cloud Generation from 2D Images in Unsupervised Reconstruction}

\author{\textbf{Hongbin Lin\textsuperscript{1,2,$\dagger$}}\\
\and
\textbf{Juangui Xu\textsuperscript{1,3,$\dagger$}}\\
\and
\textbf{Qingfeng Xu\textsuperscript{1}} \\
\and
\textbf{Zhengyu Hu\textsuperscript{2}}
\and
\textbf{Handing Xu\textsuperscript{1}}
\and
\textbf{Yunzhi Chen\textsuperscript{1}}
\and
\textbf{Yongjun Hu\textsuperscript{3}}
\and
\textbf{Zhenguo Nie\textsuperscript{1}\thanks{Corresponding author. \textsuperscript{{$\dagger$}}These authors contributed equally to this work. Preprint. Under review.}}
\and
\textsuperscript{1}Tsinghua University\quad
\textsuperscript{2}HKUST(GZ)\quad
\textsuperscript{3}Guangzhou University\\
}

\begin{document}
\maketitle
\begin{abstract}
In the realm of 3D reconstruction from 2D images, a persisting challenge is to achieve high-precision reconstructions devoid of 3D Ground Truth data reliance. We present UNeR3D, a pioneering unsupervised methodology that sets a new standard for generating detailed 3D reconstructions solely from 2D views. Our model significantly cuts down the training costs tied to supervised approaches and introduces RGB coloration to 3D point clouds, enriching the visual experience. Employing an inverse distance weighting technique for color rendering, UNeR3D ensures seamless color transitions, enhancing visual fidelity. Our model's flexible architecture supports training with any number of views, and uniquely, it is not constrained by the number of views used during training when performing reconstructions. It can infer with an arbitrary count of views during inference, offering unparalleled versatility. Additionally, the model's continuous spatial input domain allows the generation of point clouds at any desired resolution, empowering the creation of high-resolution 3D RGB point clouds. We solidify the reconstruction process with a novel multi-view geometric loss and color loss, demonstrating that our model excels with single-view inputs and beyond, thus reshaping the paradigm of unsupervised learning in 3D vision. Our contributions signal a substantial leap forward in 3D vision, offering new horizons for content creation across diverse applications. Code is available at \url{https://github.com/HongbinLin3589/UNeR3D}.
\end{abstract}
\section{Introduction}
\label{sec:intro}

\begin{figure}[!ht]
    \centering
    \includegraphics[width=0.85\linewidth]{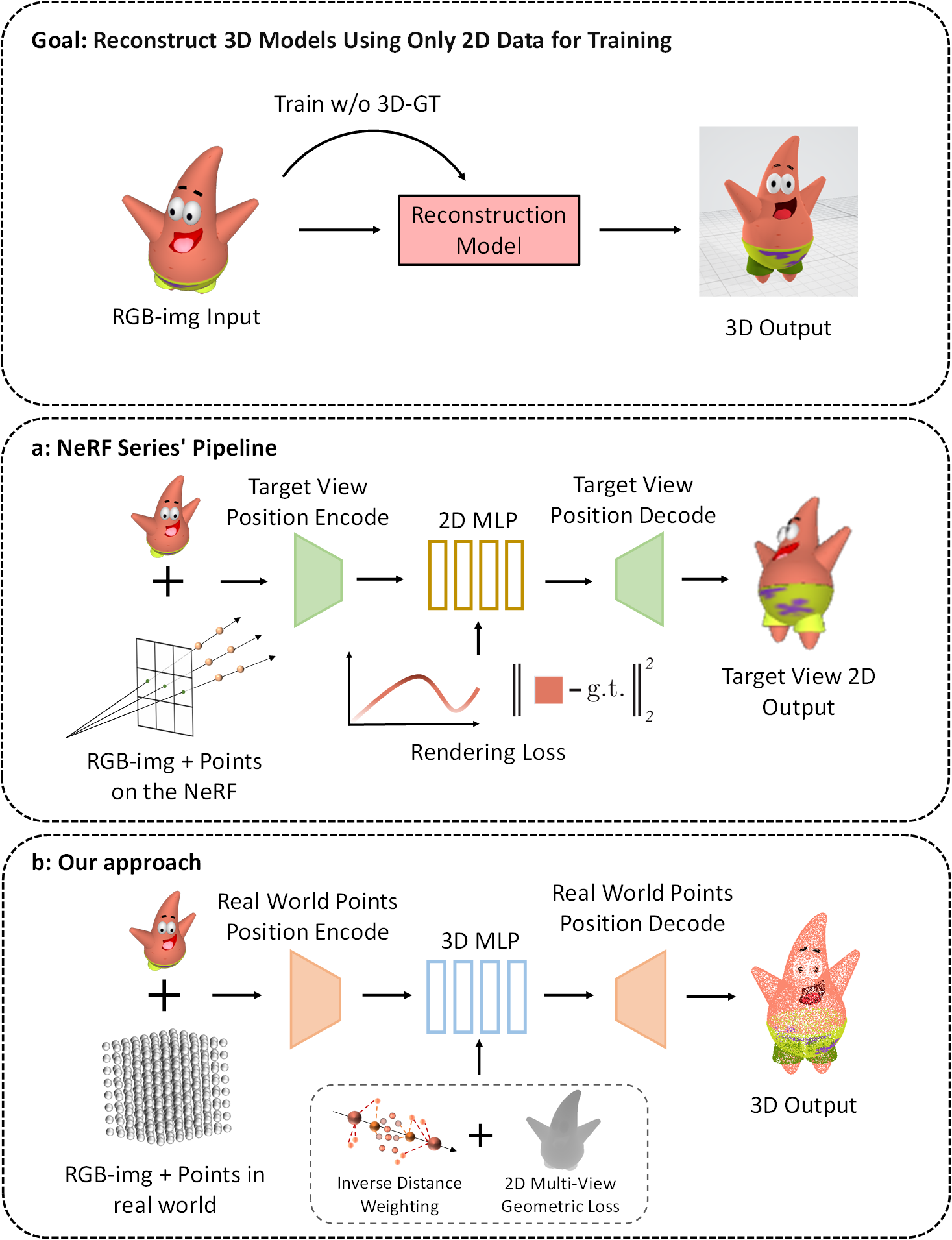}
    \caption{\textbf{Comparative Overview of 3D Reconstruction.} Our goal is to train a model for reconstructing 3D data solely from 2D images. (a) The NeRF series\cite{mildenhall2021nerf,yu2021pixelnerf,xu2022point,wang2021neus} encodes points on the neural radiance field, outputting 2D images of the target camera. (b) Our approach, by encoding average sampled points from the real world and integrating inverse distance weighting with multi-view geometric loss, reconstructs realistic 3D models.}
    \label{fig:inrto}
\end{figure}

Image-based 3D reconstruction, the process of restoring 3D information from 2D images, has been a key focus in computer vision research, driven by its applications in diverse fields such as robotics\cite{thrun2002probabilistic}, object recognition\cite{he2016deep}, and medical diagnostics\cite{litjens2017survey}. This transformation from 2D to 3D is crucial for applications where understanding 3D structures from 2D images is vital, including industrial automation and digital avatar creation\cite{hong2022avatarclip}.

Advancements in 2D image processing and computational power have significantly influenced the evolution of 3D reconstruction\cite{han2019image,zhu2023deep}. Traditional methods like Structure from Motion (SfM)\cite{schonberger2016structure}, Multi-view Stereo (MVS)\cite{furukawa2015multi}, and Photometric Stereo\cite{herbort2011introduction} have utilized multiple images and geometric constraints. The rise of deep learning has further revolutionized this field, enhancing accuracy and complexity handling through volumetric\cite{wu20153d,maturana2015voxnet}, point-based\cite{qi2017pointnet}, and mesh-based methods\cite{wang2018pixel2mesh}, along with implicit function representations for complex shape delineation\cite{sitzmann2019scene,park2019deepsdf}.

Despite these advancements, the widespread reliance on 3D Ground Truth data for training models in 3D reconstruction presents significant challenges\cite{dai2017shape}. Acquiring this data typically requires costly and sophisticated equipment like laser lidars and RGB-D scanners, leading to substantial time and computational costs\cite{dou2016fusion4d}. After acquisition, extensive processing is necessary to derive usable 3D representations\cite{berger2017survey, rusinkiewicz2001efficient, newcombe2011kinectfusion}. The relative scarcity of 3D datasets, compared to 2D, hampers the development of 3D reconstruction methods\cite{silberman2012indoor}, emphasizing the urgent need for innovative, less data-dependent reconstruction techniques.

Neural Radiance Fields (NeRF) and its variants\cite{mildenhall2021nerf,yu2021pixelnerf,ge2023ref,xu2022point,wang2021neus,wang2022hf,meng2023neat} have significantly advanced neural network-based 3D structure representations, improving complex 3D scene rendering. However, they generally lack the ability to generate explicit 3D models, encoding structures implicitly and requiring viewpoint-specific recomputation. The recent focus in generative AI on combining 2D imagery with 3D modeling has led to attempts to integrate NeRF into this domain. Yet, these efforts face challenges like pose alignment due to NeRF's inability to provide direct 3D outputs\cite{liu2023one,li2023adv3d,singer2023text,jain2022zero}.

Addressing the challenge of developing a generalizable algorithm that relies solely on 2D images for training and produces explicit 3D models, we introduce UNeR3D, an unsupervised 3D reconstruction method infused with 3D prior knowledge. UNeR3D employs ResNet34 for feature extraction and a specialized MLP with point's positional encoding, facilitating the creation of 3D RGB point clouds by combining image data with inverse distance weighting. This method, enhanced by multi-view geometric and color losses, supports various input views, including single-view, for versatile reconstruction. Experimental results demonstrate UNeR3D's superior performance and efficiency in 3D vision, highlighting its potential to transform the field. Our reconstruction approach is poised to tackle a broader range of 3D reconstruction tasks, particularly those requiring alignment of text-to-3D generation.

\textbf{Our Contributions:}
\begin{enumerate}
    \item To the best of our knowledge, UNeR3D is the inaugural model that leverages unsupervised learning to generate high-precision 3D point clouds directly from 2D imagery, substantially reducing the need for labor-intensive, supervised training methods.
    \item We introduce the first method within the NeRF framework capable of reproducing RGB attributes on point clouds, enabling the synthesis of high-fidelity images adaptable to a variable number of input views for enhanced reconstruction flexibility.
    \item Our approach employs inverse distance weighting for neural rendering, innovatively ensuring smooth color transitions. The trained model can infer point clouds at arbitrary densities, allowing for reconstructions of any desired precision.
\end{enumerate}
\section{Related work}
\subsection{Deep Learning-based 3D Reconstruction}
\label{sec:related1}
Recent years have seen an upsurge in deep learning techniques dedicated to 3D object reconstruction, prominently structured under voxel-based, mesh-based, and point cloud-based representations. While these methods have garnered impressive results, a common denominator is their reliance on supervised learning paradigms, often necessitating meticulously curated ground truth 3D data for effective training—a significant bottleneck in real-world applicability. Early forays into voxel-based representations used volumetric grids for capturing spatial outlines of objects\cite{choy20163d,girdhar2016learning}. Later methods leveraged adaptive octree structures, offering better efficiency and resolution\cite{wang2018adaptive,riegler2017octnet}, while others integrated context-awareness to bridge single and multi-view reconstructions\cite{xie2019pix2vox,wang2017shape}. Shifting gears to mesh-based reconstructions, frameworks emerged that transformed RGB images directly into 3D mesh structures\cite{wang2018pixel2mesh,pontes2019image2mesh}. The marriage of graphics with deep learning led to realistic surface renderings\cite{kato2018neural}. The point cloud domain heralded architectures for unordered point set processing\cite{qi2017pointnet,qi2017pointnet++}, and subsequent innovations targeted dense and accurate point cloud reconstructions\cite{mandikal2019dense,wang2019mvpnet}. Innovations like geometry-aware transformers ushered advancements in point cloud completion tasks\cite{yu2021pointr}. This exposition underscores a poignant lacuna in the field: the quest for a genuine unsupervised learning-based 3D reconstruction method, which we address in this work.

\begin{figure*}[!ht]
    \centering
    \includegraphics[width=0.95\linewidth]{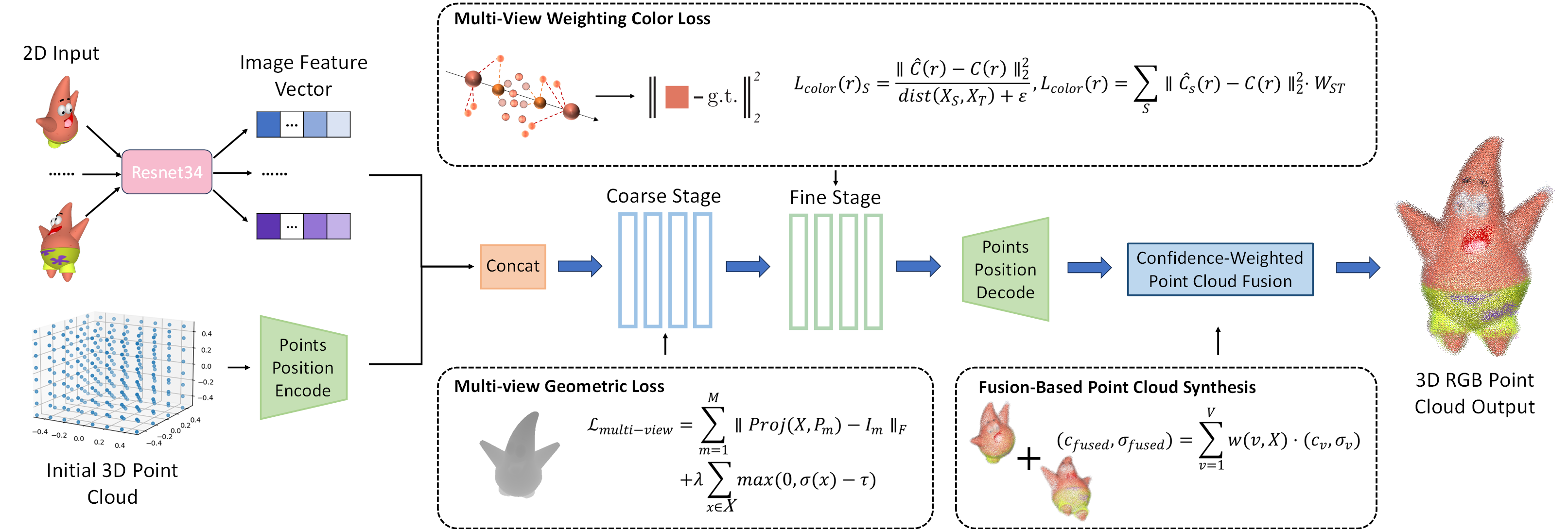}
    \caption{\textbf{UNeR3D Pipeline: Transforming 2D Images into 3D RGB Point Clouds.} The pipeline integrates feature extraction from 2D images, positional encoding of the initial point cloud, and a two-stage neural network processing. In the coarse stage, the Multi-view Geometric Loss shapes the geometric outline of the target object, which is then refined in the fine stage using the Multi-View Weighting Color Loss to endow each point in the point cloud with lifelike colors. The final 3D RGB point cloud is reconstructed by applying confidence weighting.}
    \label{fig:pipeline}
\end{figure*}

\subsection{Positional Encoding and Unsupervised Learning}
\label{sec:related2}
Positional encoding plays a pivotal role in neural rendering, particularly in enhancing models like NeRF to represent complex spatial details\cite{mildenhall2021nerf}. This technique, which transforms input coordinates for neural networks to capture high-frequency functions better, is instrumental in models' ability to learn detailed geometries, as demonstrated in seminal works such as Vaswani et al.'s\cite{vaswani2017attention}. Its application in NeRF highlights its efficacy in achieving detailed 3D scene representations without extensive supervision\cite{tancik2020fourier}, further expanding its utility, non-parametric models like Point-NN leverage non-learnable operations, allowing for sophisticated geometric analysis in point cloud data\cite{zhang2023parameter}.

Unsupervised 3D reconstruction has evolved through integrating SfM and MVS techniques, essential for deducing spatial depth from imagery\cite{schonberger2016structure,furukawa2015multi}. While effective, such methods are traditionally limited by their high computational load and the required extensive array of input views. Groundbreaking efforts, such as Yan et al.'s Perspective Transformer Nets, reduce reliance on 3D ground truth data\cite{yan2016perspective}, and Liu et al.'s CAPNet utilizes 2D data to approximate 3D point clouds\cite{navaneet2019capnet}. However, these techniques still encounter limitations from specific assumptions and prior knowledge needs. Advances in video-based 3D reconstruction by Smith et al. and targeted unsupervised learning for symmetric objects by Jackson et al. showcase the field's progress while highlighting the ongoing need for a flexible and universally applicable unsupervised 3D reconstruction method\cite{henzler2021unsupervised, wu2020unsupervised,hu2023leveraging}.

\subsection{Implicit 3D reconstruction}
\label{sec:related3}
Implicit representations have redefined the landscape of 3D reconstruction, offering precise surface delineation while minimizing the computational load when contrasted with traditional methods like occupancy grids or voxel frameworks\cite{mescheder2019occupancy}. NeRF stands at the vanguard of this innovation, enabling the generation of volumetric scenes through neural networks, which facilitate photorealistic renderings. PixelNeRF has propelled the adaptability of NeRF by incorporating feature encodings from input images into the MLP, endowing it with the ability to generalize across different scenes\cite{yu2021pixelnerf}. On a parallel front, Point-NeRF\cite{xu2022point} merges deep multi-view stereo with NeRF to refine rendering, using point clouds to direct the process for enhanced precision and efficiency. Despite its improvements in visual quality, it does not generate point clouds through NeRF's structure and lacks generalizability, fundamentally remaining a 2D rendering model. However, these methods, including advanced variants like NeuS\cite{wang2021neus}, HF-NeuS\cite{wang2022hf}, and NeAT\cite{meng2023neat}, still largely remain within the domain of implicit reconstruction, not directly yielding tangible 3D models from the NeRF framework itself. This limitation has implications for multi-modal alignment in text-to-3D applications, as current NeRF-based methods do not inherently provide a generalizable framework for explicit 3D reconstruction. This gap highlights the need for new approaches to produce usable and adaptable 3D models from 2D inputs directly.

\section{Methodology}
We present a novel strategy for unsupervised 3D reconstruction from 2D imagery by integrating the neural radiance field concept with a knn-based inverse distance weighting scheme (see Fig.~\ref{fig:pipeline}). Our method computes the RGB color and confidence of presence for 3D points directly from their coordinates and the associated camera pose of the input image. The reconstruction process consists of querying the RGB color and confidence for each point in the initial point cloud and constructing a refined point cloud that accurately reflects the object's true 3D structure.

The reconstructed point cloud $S$ is formed by aggregating points that surpass a predefined confidence threshold $\theta$:
\begin{equation}
S = \bigcup_{\mathbf{x} \in Q} \{ f(\mathbf{I}, \mathbf{P}, \mathbf{x}) | \alpha(\mathbf{x}) > \theta \}
\end{equation}
Here, $Q$ denotes the initial point cloud, and the function $f$ determines the RGB color and the existence confidence of a point $\mathbf{x}$ within this point cloud. And $\mathbf{P}$ represents the camera pose of the input image $\mathbf{I}$.

Function $f$ is subsequently defined as:
\begin{equation}
f(\mathbf{I}, \mathbf{P}, \mathbf{x}) = (\mathbf{c}, \alpha)
\end{equation}
where $\mathbf{c}$ represents the predicted RGB color and $\alpha$ signifies the confidence level of the point's existence.

In our framework, the alignment of the generated point cloud with the input imagery is facilitated by a composite loss function:
\begin{equation}
\mathcal{L} = \mathcal{L}_{geom}(S, \mathbf{I}) + \lambda \mathcal{L}_{color}
\end{equation}
Here, $\mathcal{L}_{geom}$ represents the multi-view geometric loss ensuring structural consistency across different views, and $\mathcal{L}_{color}$ enforces color consistency. The parameter $\lambda$ serves as a balancing coefficient between the two losses. 

Subsequent subsections will delve into each component of our methodology, providing a detailed description of the entire pipeline from initial feature extraction to the final 3D structure colorization.

\subsection{Initial Point Cloud Generation and Color Discrimination}
We initiate our reconstruction process by formulating an initial point cloud $\text{X} = S(\text{r})$ from a view-camera pose pair $(\text{I}, \text{P})$. The resolution superparameter $\text{r}$ governs the granularity of this cubic lattice. This cloud is then transformed from the world to the camera coordinate system using the camera pose $\text{P}$.

Positional encoding of points within $\textbf{X}$ leverages sinusoidal functions to facilitate a high-dimensional representation conducive to neural network learning:

\begin{multline}
\text{PosE}(p_i)=\text{Concat}(\sin(\alpha p_i+\beta^{j/C}), \cos(\alpha p_i+\beta^{j/C})),\\
\ j \in [0, C/6),
\end{multline}
where $C$ is a multiple of 6, indicative of the encoding's dimensionality\cite{zhang2023parameter}.

We employ ResNet-34 to extract feature maps $\text{F} = E(\text{I})$ from $\text{I}$. This step is pivotal for the model's generalization across diverse data. Projecting points in $\text{X}$ onto $\text{F}$ captures view-dependent features, which, alongside positional encodings and the direction vector $\text{PosE}(\text{d})$, inform the MLP to predict volume density $\sigma$ and RGB values $\text{rgb}$:

\begin{equation}
(\sigma, \text{rgb}) = N(\pi(\text{F}, \text{P}, \text{X}), \text{PosE}(\text{X}),\text{PosE}(\text{d}))
\end{equation}

We construct a 3D representation where points with density $\sigma$ exceeding the threshold $\theta$ (set to zero) are retained and colored, forming a detailed point cloud of the observed object. Using the source view as the coordinate system stems from our goal to generate a 3D model corresponding to world coordinates, which can be directly transformed into any target view without additional computation. This approach enhances the efficiency and versatility of the reconstruction process, allowing for seamless integration into various application scenarios.

\subsection{Coarse-to-Fine Stage Structure}
\label{sec:coarse_fine_structure}
Our methodology employs a coarse-to-fine structure to process the point cloud, enhancing the reconstruction's overall quality by focusing on different objectives at each stage\cite{mildenhall2021nerf,yu2021pixelnerf}. The specific loss functions utilized in these stages are detailed in Section \ref{sec:loss_function}.

The coarse stage centers on reconstructing the geometric outline of the target object. Sampling points from the initial point cloud focus on establishing the basic geometric shape, laying a foundation for the detailed reconstruction process. Following the coarse stage, the fine stage refines the details and colors of the model. It processes a denser point cloud sample, concentrating on intricate aspects and color accuracy, which is crucial for the final high-quality 3D model.

This structured approach enables our model to establish the basic shape and structure in the coarse stage and then add detailed refinements in the fine stage, resulting in an effective and high-quality 3D reconstruction.

\subsection{Neural Rendering with Inverse Distance Weighting}
\label{subsec:neural_rendering_idw}
Considering that actual light behaves not as a singular ray but rather as a viewing cone with a projection area\cite{barron2021mip}, our methodology augments neural radiance rendering by integrating an inverse distance weighting strategy. This approach synthesizes point positions and colors, diverging from the traditional NeRF's 2D rendering limitations. The normalized weighting scheme is critical for supervising the structural and color accuracy of the 3D point cloud, thereby enhancing detail in the reconstructed model.

\begin{figure}[!ht]
    \centering
    \includegraphics[width=0.45\linewidth]{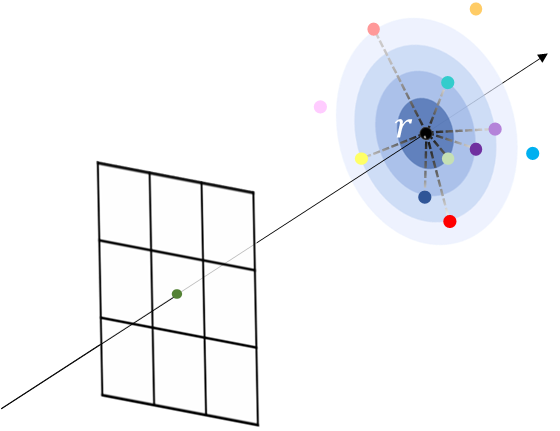}
    \caption{\textbf{Inverse Distance Weighting in UNeR3D's Rendering for Realistic Coloration.} Our approach posits that points further from the neural radiance line influence color rendering less. Our inverse distance weighting captures this, where each point's color impact is inversely related to its distance to the radiance line. This method prioritizes points closer to the object's surface, ensuring a realistic coloration of the point cloud and enhancing the visual authenticity of the reconstructed models.}
    \label{fig:IDW}
\end{figure}

Figure \ref{fig:IDW} illustrates our approach where the neural point voxel density-color pair $(\sigma_r,\text{c}_r)$ for each point is calculated by aggregating the $k$-nearest neighbors within a radius $r$:

\begin{equation}
\label{eq:inverse_weighting}
(\sigma_r, \text{c}_r) = \left(\frac{ \sum_{i=1}^{k} w_i \sigma_i}{\sum_{i=1}^{k} w_i+\varepsilon } , \frac{\sum_{i=1}^{k} w_i \sigma_i c_i}{\sum_{i=1}^{k} w_i \sigma_i+\varepsilon}  \right),
\end{equation}

\noindent The weights $w_i$ are normalized to allow for proportional contributions of neighboring points, calculated as:

\begin{equation}
\label{equ:dis_weights}
w_i = \frac{1}{\text{dist}(r, p_i)^2 + \varepsilon},
\end{equation}

\noindent where $\text{dist}(r, p_i)$ is the Euclidean distance between the radiance point $r$ and a point $p_i$ in the point cloud, with $\varepsilon$ preventing division by zero. By focusing on points with non-negligible density $\sigma_i$, we mitigate the influence of noise points that do not represent the object, ensuring they do not adversely affect the color rendering.

By employing this inverse distance weighting, our rendering protocol guarantees smoother color transitions within a specified neighborhood, avoiding abrupt changes that could disrupt the visual realism. The refinements to the number of points $k$ and the radius $r$ are instrumental in elevating the accuracy and density of the point cloud, significantly contributing to the advancement of unsupervised 3D RGB point cloud reconstruction.

\subsection{Loss Function Design}

\begin{figure*}[!ht]
    \centering
    \includegraphics[width=0.9\linewidth]{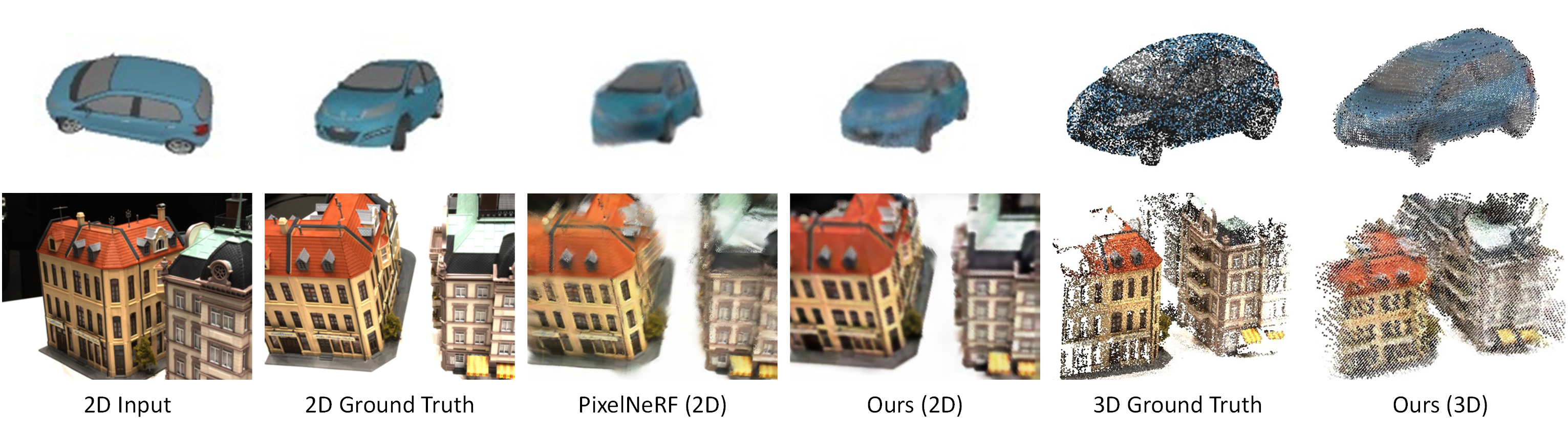}
    \caption{Comparative Generalization Performance on Unseen Images from ShapeNet and DTU Datasets: A side-by-side comparison between UNeR3D and PixelNeRF demonstrates the enhanced clarity and detail resolution achieved by our method, underlining its superior ability in generalizing to unseen images with greater visual fidelity.}
    \label{fig:compare2d}
\end{figure*}

\label{sec:loss_function}
\noindent\textbf{Multi-view Geometric Loss}
During the coarse stage of training, our model emphasizes the alignment of global structure with the application of a multi-view geometric loss. This loss component is pivotal for establishing the foundational geometry of the reconstructed point cloud across multiple views. By leveraging the geometric adversarial loss (GAL) principles\cite{jiang2018gal}, we utilize known camera poses to project the point cloud onto image planes, aligning it with the silhouettes captured in the corresponding images.

With camera pose matrices $\{\text{P}_m\}_{m=1}^M$ and corresponding images $\{\text{I}_m\}_{m=1}^M$, the geometric loss is computed as:

\begin{equation}
\mathcal{L}_{\text{geom}} = \sum_{m=1}^M \|\text{Proj}(\textbf{X}, \text{P}_m) - \text{I}_m\|_F,
\end{equation}

\noindent where $\text{Proj}(\cdot)$ projects points in the point cloud $\textbf{X}$ onto the image plane, and $\|\cdot\|_F$ represents the Frobenius norm.

To distinguish points related to the object from background noise, we introduce a regularization term:

\begin{equation}
\mathcal{L}_{\text{reg}} = \lambda \sum_{x \in \textbf{X}} \max(0, \sigma(x) - \tau),
\end{equation}

with $\lambda$ as the regularization weight and $\tau$ as the threshold for object relevance.

The coarse stage loss thus combines projection accuracy and regularization:

\begin{equation}
\mathcal{L}_{\text{coarse}} = \mathcal{L}_{\text{geom}} + \mathcal{L}_{\text{reg}},
\end{equation}

optimizing for a coherent geometric structure that serves as a scaffold for further refinement in subsequent stages. This multi-view geometric consistency is crucial for the early stages of network training, ensuring the model's global understanding of the 3D structure aligns with the varied perspectives offered by the input views.

\noindent\textbf{Color Consistency with Multi-View Weighting}
Our color consistency loss function, particularly pertinent during the fine stage of our model's training, integrates multi-view considerations to ensure accurate color reproduction. The loss for a ray $r$, formulated as follows, weighs the contributions from various source views against the target view utilized during the supervised training phase:

\begin{equation}
\mathcal{L}_{\text{color}}(r)_{S} = \frac{\|\hat{c}(r) - C(r)\|_{2}^{2}}{\text{dist}(X_{S}, X_{T}) + \varepsilon},
\end{equation}

\noindent where $\hat{c}(r)$ represents the estimated color along ray $r$, $C(r)$ is the actual color from the ground truth, and $\text{dist}(X_{S}, X_{T})$ is the distance between the source view $X_{S}$ and the target view $X_{T}$ used for training supervision. The addition of a small constant $\varepsilon$ prevents division by zero.

The aggregated color loss across all views is then given by:

\begin{equation}
\mathcal{L}_{\text{fine}} = \mathcal{L}_{\text{color}}(r) = \sum_{S} \|\hat{C}_{s}(r) - C(r)\|_{2}^{2} \cdot W_{ST},
\end{equation}

\noindent with the view weighting factor $W_{ST}$ normalized across all source views $S$ relative to the target view $T$:

\begin{equation}
W_{ST} = \frac{\frac{1}{\text{dist}(x_{S}, x_{T}) + \varepsilon}}{\sum_{S} \frac{1}{\text{dist}(x_{S}, x_{T}) + \varepsilon}}.
\end{equation}

This loss function ensures that the computed colors are consistent with the observed ones, effectively training the model to render accurate and coherent images across multiple viewpoints.

\subsection{Fusion-Based Point Cloud Synthesis}
\label{sec:fused_synthesis}
Our model's architecture is adept at handling a flexible number of views for reconstruction, from a single view to multiple views. This flexibility is achieved through a unified weighting scheme that adapts to the available input views for single-view or multi-view reconstructions.

The synthesis of the point cloud is directed by an equation that uniformly accounts for the contribution of each view:

\begin{equation}
\label{eq:fusion_synthesis}
(c_{\text{fused}}, \sigma_{\text{fused}}) = \sum_{v=1}^{V} w(v, X) \cdot (c_v, \sigma_v),
\end{equation}

\noindent Here, $V$ denotes the number of views available for reconstruction ($V=1$ signifies a single-view), $X$ represents a point within the point cloud, and $w(v, X)$ is the calculated weight for view $v$ at point $X$. The weights are derived by combining a distance-based factor and the transmittance along the ray from each view:

\begin{equation}
w(v, X) = \frac{W_{k}(v, X)}{\max \{W_{k}(v, X)\}} \exp \left(-\int_{s_{n}}^{s} \sigma(\mathbf{r}_v(s)) \, ds\right),
\end{equation}

\noindent where $W_{k}(v, X)$ is the distance-based weight for view $v$ and point $X$, as defined by Equation \ref{equ:dis_weights}. This normalization ensures that each view's contribution is proportionally adjusted based on its distance to the neural radiance point, with closer views having a more significant influence on the final point cloud.

Our method enables reconstruction from any number of views post-training, including a single image, but also facilitates the derivation of point clouds at any desired density, allowing for reconstructions of varying precision that cater to different resolution requirements without additional training. The synthesis process, which adaptively weights the contributions of each view, accentuates the model's versatility by leveraging the MLP's learned priors along with the geometric and photometric data from input views, enabling the generation of detailed and precise 3D point clouds, showcasing our model's ability to integrate and adaptively apply comprehensive learned information to a wide range of reconstruction scenarios.
\section{Experiments}
In this section, we present a comprehensive evaluation of our model, contrasting it against state-of-the-art NeRF variants including PixelNeRF\cite{yu2021pixelnerf}, PointNeRF\cite{xu2022point}, and NeuS\cite{wang2021neus}, using the ShapeNet\cite{wu20153d} and DTU\cite{aanaes2016large} datasets. Furthermore, we conduct a series of ablation studies to dissect the contributions of each component within our framework, substantiating the superiority of our proposed model.

\begin{table*}[htbp]
\centering
\caption{Performance Comparison on ShapeNet and DTU datasets}
\label{tab:performance_comparison}
\begin{tabular}{lcccc}
\toprule
Metric & UNeR3D (Ours) & PixelNeRF\cite{yu2021pixelnerf} & PointNeRF\cite{xu2022point} & NeUS\cite{wang2021neus} \\
\midrule
2D SSIM↑ (Generalization) - ShapeNet & \textbf{0.957} & 0.918 & - & - \\
2D SSIM↑ (Generalization) - DTU & \textbf{0.908} &0.823  & - & - \\
2D PSNR↑ (Generalization) - ShapeNet & \textbf{28.413} & 22.783 & - & - \\
2D PSNR↑ (Generalization) - DTU & \textbf{25.617} &16.634  & - & - \\
2D PSNR↑ (Single Scene) - DTU & 28.468 & - & 21.367 & \textbf{30.851} \\
2D SSIM↑ (Single Scene) - DTU & 0.873 & - & 0.818 & \textbf{0.931}\\
3D EMD↓ (Single Scene) - DTU & \textbf{0.362} & - & 0.672 &  0.813\\
\bottomrule
\end{tabular}
\end{table*}

\begin{figure*}[!ht]
    \centering
    \includegraphics[width=0.95\linewidth]{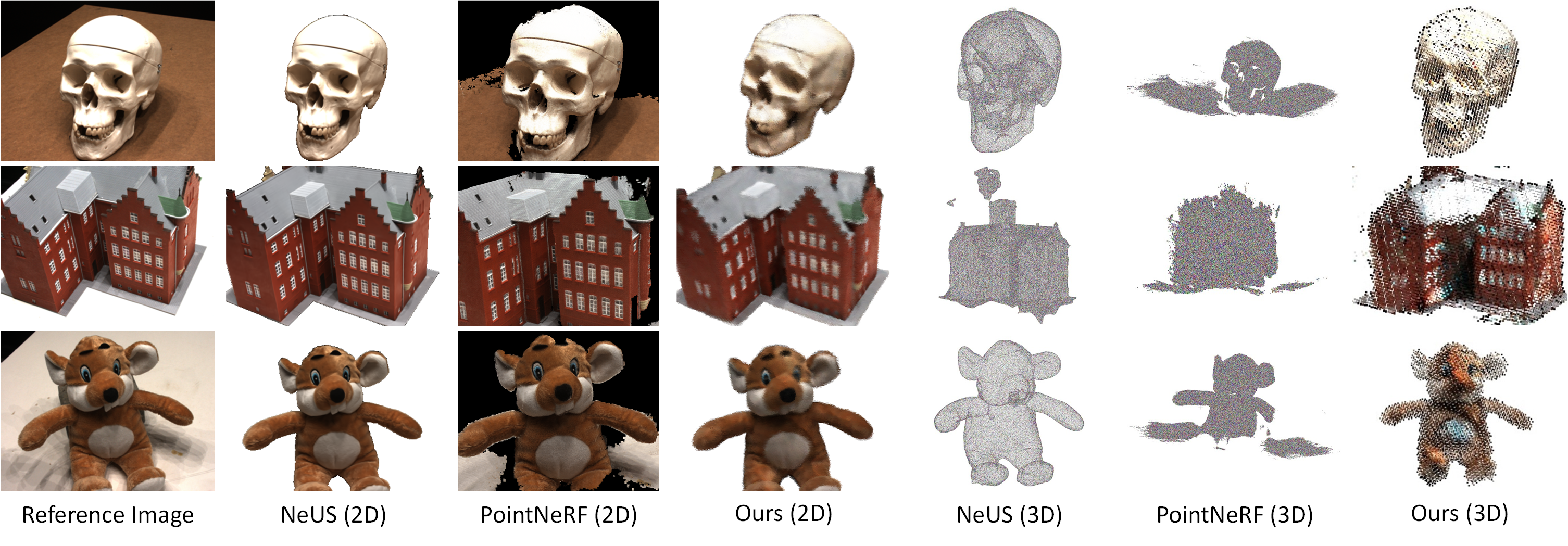}
    \caption{Comparison of 2D and 3D reconstruction outcomes on a single scene from the DTU dataset: UNeR3D versus PointNeRF and NeUS. Our approach, with its unique capability to restore color in 3D reconstruction, achieves more lifelike and visually compelling 3D results.}
    \label{fig:compare3d}
\end{figure*}

\subsection{Datasets}
\noindent\textbf{Shapenet}
ShapeNet is an extensive collection of 3D objects with rich annotations\cite{wu20153d}. It covers a diverse set of categories, including furniture, vehicles, and electronics, among others, amounting to over 50,000 unique 3D models. Each category comprises a substantial number of examples, which supports training robust models capable of generalizing across various shapes and structures.

\noindent\textbf{DTU}
The DTU dataset, on the other hand, is a photogrammetric dataset containing images of intricate 3D scenes, designed specifically for 3D reconstruction tasks\cite{aanaes2016large}. It provides high-resolution images and accurate 3D surface models, contributing to a total of several scenes with different textures, geometries, and lighting conditions.

\subsection{Experimental Setup}
\noindent\textbf{Implementation Details}
In our implementation, the camera's rotation coordinates were set within a spherical coordinate system, with the longitude and latitude angles ranging from $0$ to $2\pi$, allowing for complete environmental capture. The camera's translation coordinates, and the point cloud coordinate system were defined in a three-dimensional Cartesian coordinate system. To facilitate calculations and maintain consistency, the range for the point cloud Cartesian coordinates was normalized to $[-1, 1]$ across all three axes. Our training protocol included different numbers of views per object—2, 4, and 8—to investigate the impact of view count on the model's ability to assimilate and generalize prior knowledge. The division of our datasets into training, validation, and test sets adhered to a 7:2:1 ratio, ensuring a thorough assessment across various sections of the data.

\noindent\textbf{Evaluation Metrics}
We employed the PSNR and SSIM\cite{wang2004image} for RGB image fidelity, and the Earth Mover's Distance (EMD)\cite{rubner2000earth} to assess 3D point cloud reconstruction, comparing all against ground truth data. Notably, our model's training did not involve 3D data, with 3D GT only used for post-training evaluation to confirm the model's 3D inference accuracy. Notably, during training, our model was not exposed to any 3D information. The utilization of 3D GT data was strictly confined to the performance evaluation stage, ensuring an unbiased validation of the model's capability to infer 3D structures. This approach underlines the effectiveness of our model in learning from 2D data while being capable of accurate 3D predictions, a key aspect of our experimental validation.

\subsection{Comparisons}
\noindent\textbf{Baseline}
In this section, we benchmark our UNeR3D model against prominent NeRF variants, namely PixelNeRF\cite{yu2021pixelnerf}, PointNeRF\cite{xu2022point}, and NeUS\cite{wang2021neus}. PixelNeRF is notable for its ability to generate 2D images from novel viewpoints using 2D images as input, showcasing significant generalization capabilities. PointNeRF leverages MVS-generated point clouds\cite{yao2018mvsnet} to guide NeRF for refined rendering, though it lacks 3D output capabilities. NeUS excels in producing detailed 3D models through extensive iterations over a single scene but needs more generalizability. Table \ref{tab:comparison} delineates the performance comparison across various dimensions: 2D output, 3D output, 3D RGB attributes, generalization abilities, and number of training iterations, highlighting the comprehensive superiority of our approach.

\begin{table}[ht]
\centering
\caption{Comparative Results between UNeR3D (Ours) and Baseline Models}
\begin{tabular}{p{2.75cm}<{\raggedright} p{0.9cm}<{\centering} p{0.9cm}<{\centering} p{0.9cm}<{\centering} p{0.9cm}<{\centering}}
\hline
\textbf{Comparison Aspect} & \textbf{Ours} & \textbf{Pixel-NeRF} & \textbf{Point-NeRF} & \textbf{NeUS} \\
\hline
2D Output                & \checkmark & \checkmark & \checkmark & \checkmark \\
3D Output                & \checkmark & -          & \checkmark & \checkmark \\
3D RGB Attributes        & \checkmark & -          & -          & -          \\
Generalizability         & \checkmark & \checkmark & -          & -          \\
\hline
\end{tabular}
\footnotesize{\\*Note: The 3D output for PointNeRF is derived from MVS\cite{yao2018mvsnet}, not directly from the NeRF pipeline, while UNeR3D's is integrated within the NeRF pipeline.}
\label{tab:comparison}
\end{table}

\noindent\textbf{Qualitative Results}
Figure \ref{fig:compare2d} and \ref{fig:compare3d} presents a visual comparison of our UNeR3D model against the baseline models mentioned previously: PixelNeRF, PointNeRF, and NeUS. Visually, our model generates three-dimensional reconstructions with superior clarity and detail. In terms of two-dimensional imagery, our results are comparable and, in some aspects, indistinguishable from those generated by the baselines, highlighting the robustness of our approach while maintaining high fidelity in 3D structure generation.

\noindent\textbf{Quantitative Results}
We conducted training and evaluation on identical datasets and splits for our UNeR3D model and the comparison models to assess generalization capabilities. Specifically, UNeR3D demonstrated superior performance in single-scene training scenarios, achieving results comparable to NeUS with significantly fewer iterations. Our model reached a performance level similar to NeUS's 150,000 iterations at around 30,000 iterations. However, as NeUS continually converges with more iterations due to its non-generalizable nature, UNeR3D's precision plateaus beyond a certain iteration count, reflecting its inherent generalizability. Detailed quantitative outcomes are illustrated in Table \ref{tab:performance_comparison}.

\begin{figure}[!ht]
    \centering
    \includegraphics[width=0.9\linewidth]{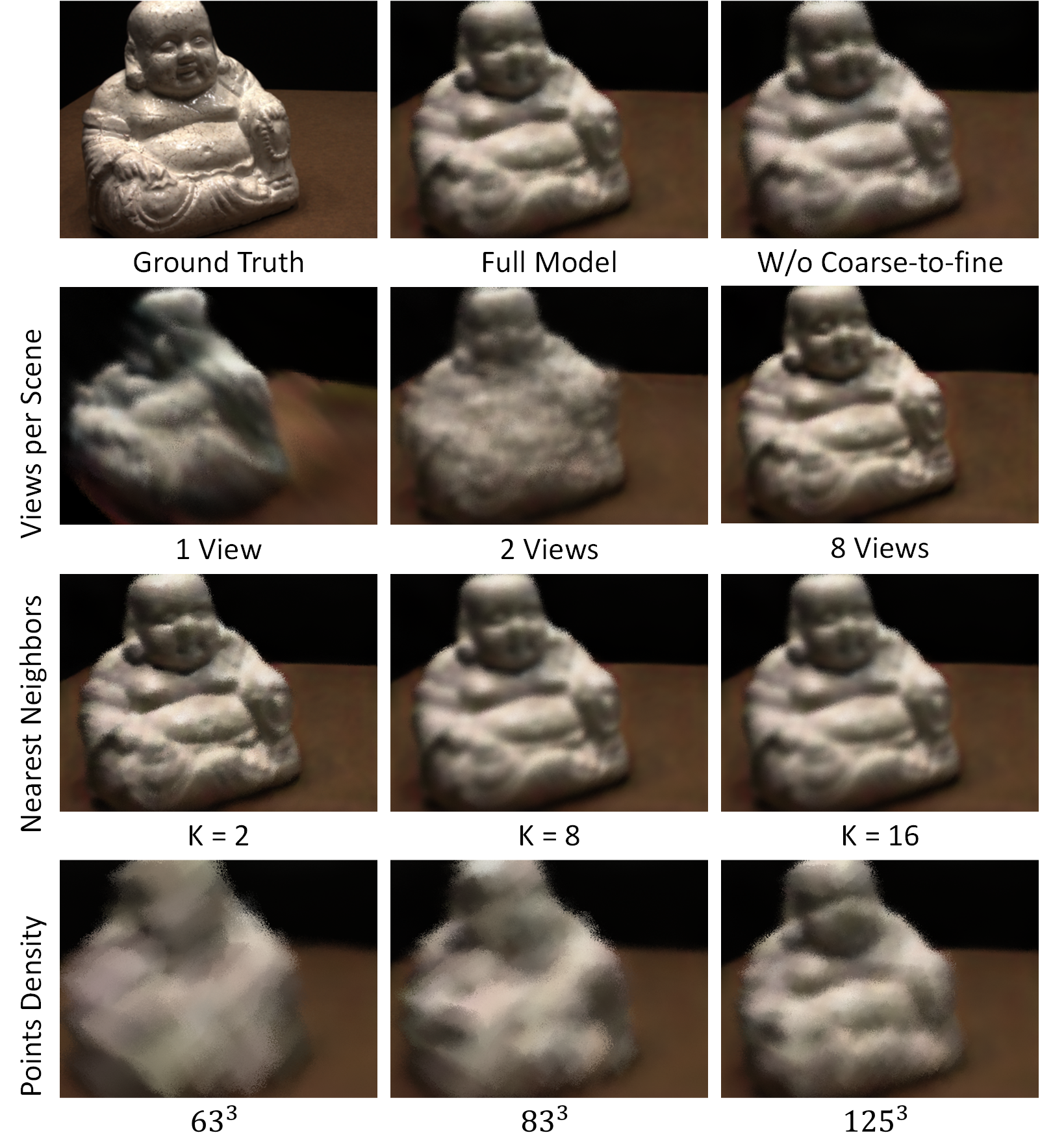}
    \caption{Comparative visualization of the ablation study results.}
    \label{fig:abstudy}
\end{figure}

\subsection{Ablation Study}
\label{sec:ablation_study}
To discern the influence of individual components within our model, we conducted an ablation study focusing on the coarse-to-fine network, the number of views per object per epoch during training, the number of nearest neighbors in the inverse distance weighting, and the initial point cloud density. We evaluated each variation on the DTU dataset, with performance measured by PSNR, SSIM, and EMD metrics. The detailed results are tabulated in Table \ref{tab:ablation_study}. Visual results of the ablation study are illustrated in Figure \ref{fig:abstudy}, and a thorough analysis with additional visualizations and discussions is included in the Appendix.

\begin{table}[h]
\centering
\caption{Ablation study results on the ShapeNet dataset, showcasing the influence of different model configurations on the PSNR, SSIM, and EMD metrics.}
\label{tab:ablation_study}
\begin{tabular}{lccc}
\hline
Configuration & PSNR↑ & SSIM↑ & EMD↓ \\ \hline
Baseline (Full Model) & \textbf{25.617} & 0.908 & \textbf{0.362} \\
w/o Coarse-to-Fine & 25.282 & 0.899 & 0.0397 \\
Views per Scene: 1 & 16.992 & 0.585 & 1.15 \\
Views per Scene: 2 & 20.325 & 0.703 & 1.10 \\
Views per Scene: 3 & 23.729 & 0.798 & 0.664 \\
Nearest Neighbors: 2 & 24.806 & \textbf{0.931} & 0.388 \\
Nearest Neighbors: 8 & 25.004 & 0.885 & 0.415 \\
Nearest Neighbors: 16 & 25.272 & 0.905 & 0.401 \\
Point Cloud Density: \(63^3\) & 19.335 & 0.717 & 0.102 \\
Point Cloud Density: \(83^3\) & 19.545 & 0.725 & 0.099 \\
Point Cloud Density: \(125^3\) & 22.137 & 0.804 & 0.687 \\ \hline

\end{tabular}
\end{table}

\section{Limitations and Conclusion}
\noindent \textbf{Limitations} Despite achieving promising results, UNeR3D faces certain limitations. Our generated 2D views still exhibit artifacts due to the model's generalizability. Furthermore, in complex scenes, the distinction between foreground and background elements can become blurred, affecting the clarity of scene structure.

\noindent \textbf{Conclusion} Our study presents UNeR3D as an innovative approach to unsupervised 3D reconstruction, effectively bypassing the need for 3D ground truth data throughout the training process. By solely utilizing 2D images, our model achieves the generation of high-fidelity 3D RGB point clouds. This achievement aligns with advances in generative AI and the shift towards creating 3D content directly from textual and 2D visual inputs. Our framework's flexibility, allowing for reconstructions from an arbitrary number of views, including single-view inputs post-training, significantly lowers the threshold for engaging in 3D reconstruction tasks and democratizes the process. As we progress in the era of generative content creation, UNeR3D holds the promise of revolutionizing 3D modeling and rendering, bridging the gap between 2D imagery and 3D digital objects. Its implications could enrich multiple fields, from virtual reality environments to robotic perception\cite{jiang2023robotic,goo2020advanced}. Our work lays a new foundation for unsupervised learning in 3D vision. It opens up exciting avenues for future research and applications in the rapidly evolving landscape of AI-driven content generation.

\section{Acknowledgement}
This work is supported by the National Natural Science Foundation of China under Grant 52175237 and the National Key Research and Development Program of China under Grant 2022YFB4703000.

{
    \small
    \bibliographystyle{ieeenat_fullname}
    \bibliography{main}
}

\clearpage
\setcounter{page}{1}
\maketitlesupplementary

\section{Additional Experimental Results}
\label{sec:appendix_additional_results}

This appendix section delves into comprehensive experimental analyses and additional results that extend beyond the scope of the main paper. These supplementary experiments aim to validate further the robustness and generalizability of our UNeR3D model across diverse datasets and real-world scenarios.

\subsection{Generalization on ShapeNet Subsets and DTU Dataset}
\begin{figure*}[!ht]
    \centering
    \includegraphics[width=1\linewidth]{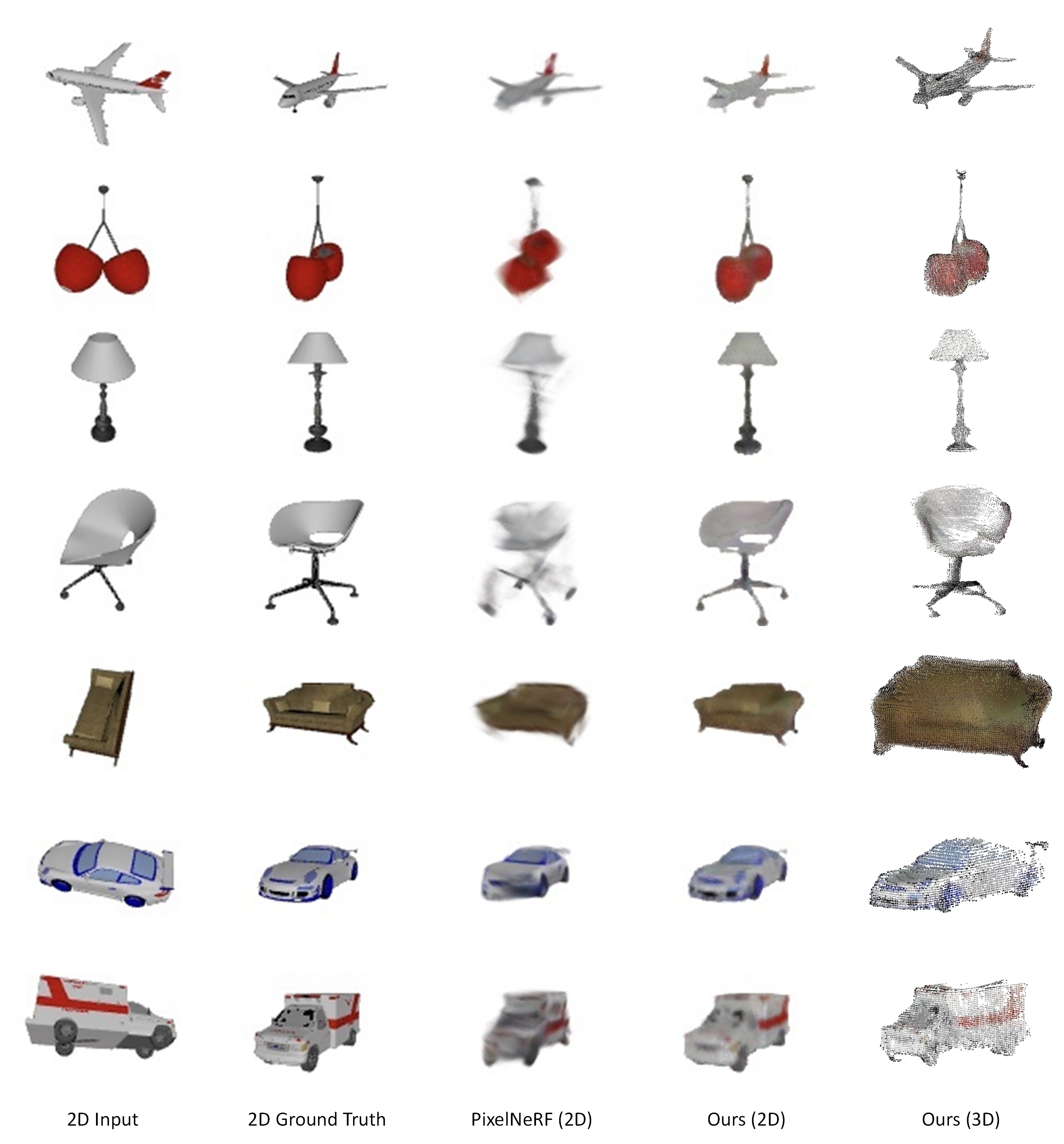}
    \caption{Comparison of 2D and 3D reconstruction outcomes on a single scene from the DTU dataset: UNeR3D versus PointNeRF and NeUS. Our approach, with its unique capability to restore color in 3D reconstruction, achieves more lifelike and visually compelling 3D results.}
    \label{fig:shapenet}
\end{figure*}

\begin{figure*}[!ht]
    \centering
    \includegraphics[width=1\linewidth]{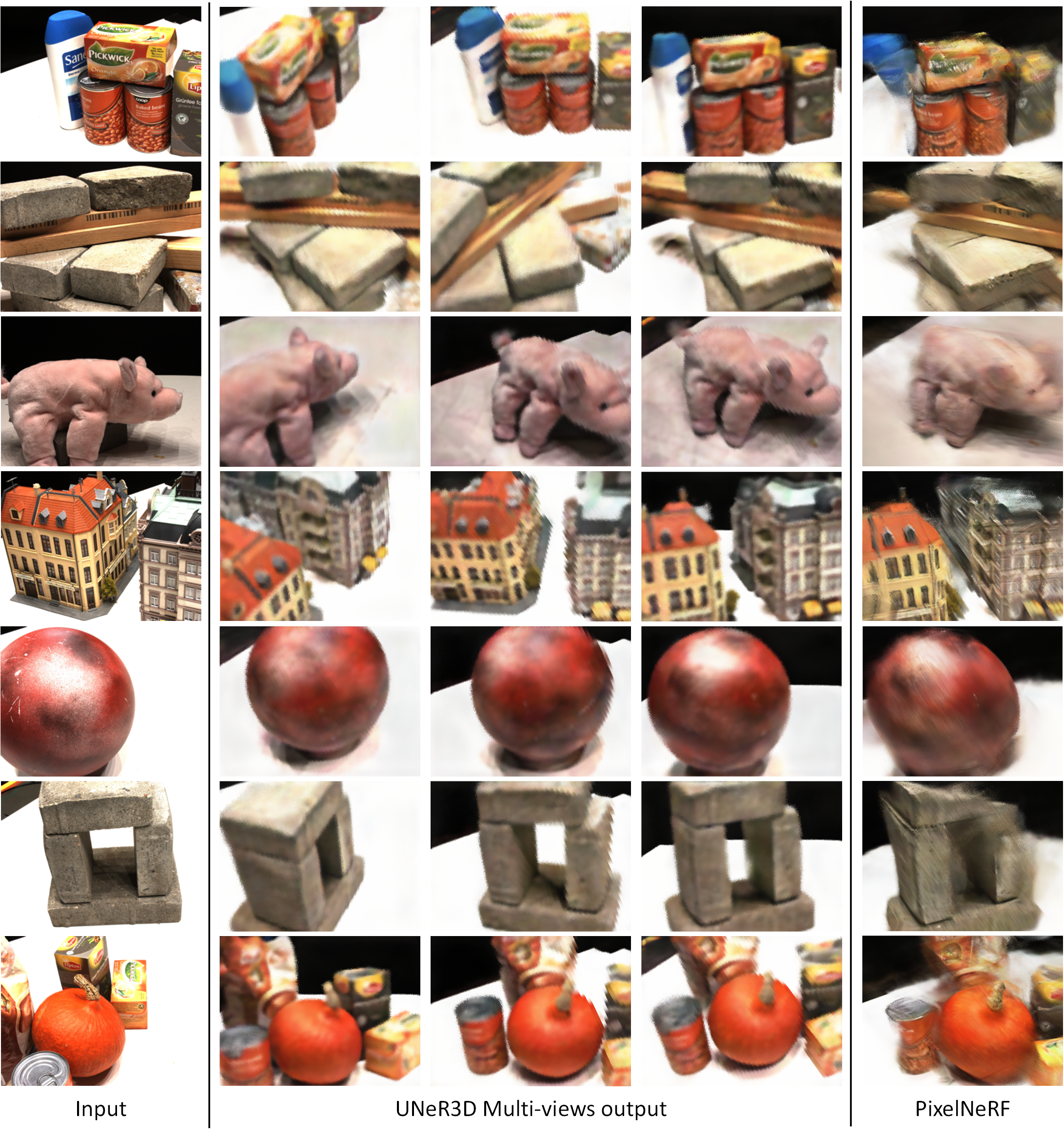}
    \caption{Demonstrating robust generalization in complex scenes, our model effectively maintains the structural integrity of objects in the DTU dataset. It showcases a clear 2D output while preserving detailed geometric features, highlighting our approach's adaptability to intricate and diverse scene compositions.}
    \label{fig:DTU}
\end{figure*}

To rigorously evaluate the generalization capabilities of our UNeR3D model, we extended our experiments to include four diverse subsets from the ShapeNet dataset: chairs, cars, lamps, and planes. These subsets were selected for their varying geometric shapes and textural complexities, posing a broad spectrum of challenges ideal for benchmarking our model's performance in an unsupervised setting. The results, depicted in Figure \ref{fig:shapenet}, highlight our model's adeptness in adapting to different object categories. This demonstrates UNeR3D's robustness in processing various 3D structures and reinforces its potential in versatile real-world applications. Furthermore, our model exhibited strong generalization performance on the DTU dataset, where PixelNeRF typically shows limitations. This underscores the enhanced adaptability of UNeR3D to various real-world scenarios. The visuals demonstrating this superior performance are presented in Figure \ref{fig:DTU}. Additionally, our ablation study, results of which are showcased in Figure \ref{fig:abstudy}, confirms the impact of each module on the model's performance, underscoring the effectiveness of our model design.

\subsection{Real-world Data Validation}
\begin{figure*}[!ht]
    \centering
    \includegraphics[width=1\linewidth]{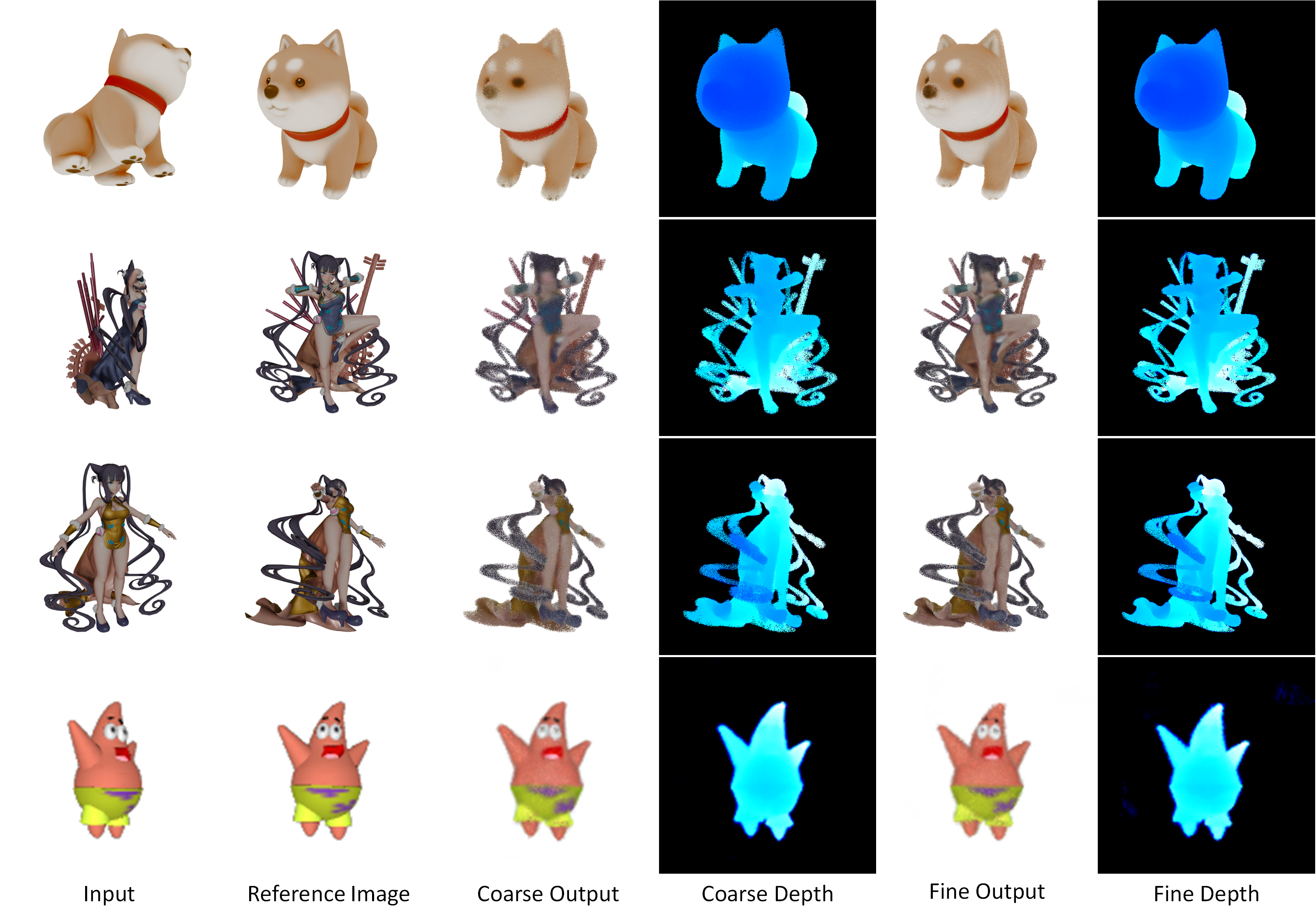}
    \caption{Reconstruction Performance of UNeR3D on Real-World Anime Character Dataset: This visualization includes images from both the Coarse and Fine stages of our model, along with their corresponding depth maps. The reconstruction showcases exceptional detail in the character model, with smooth and continuous depth variations. The colors are vivid and realistic, highlighting the effectiveness of our model in handling complex real-world data with intricate details and varied textures.}
    \label{fig:real}
\end{figure*}

In addition to standard datasets, we sought to challenge our model with real-world data, often presenting a higher degree of complexity and variability. To this end, we curated a collection of intricate models from the internet, including cartoon characters and abstract objects like 'Patrick Star' from popular media. These models, characterized by their non-standard shapes and unique textural patterns, were used to evaluate how well our model performs under real-world conditions. As depicted in Figure \ref{fig:real}, the results demonstrate our model's effectiveness in rendering realistic and coherent 3D reconstructions from diverse and complex real-world data sources.

These additional experiments underscore the versatility of UNeR3D, highlighting its potential for wide-ranging applications in 3D vision and beyond. Our findings corroborate the model's ability to generalize across different datasets, maintain high precision in detail-rich reconstructions, and adapt effectively to real-world data challenges.

\section{Network Architectures and Parameters}
\label{sec:appendix_network_architecture}

\begin{figure}[htp]
    \centering
    \includegraphics[width=1\linewidth]{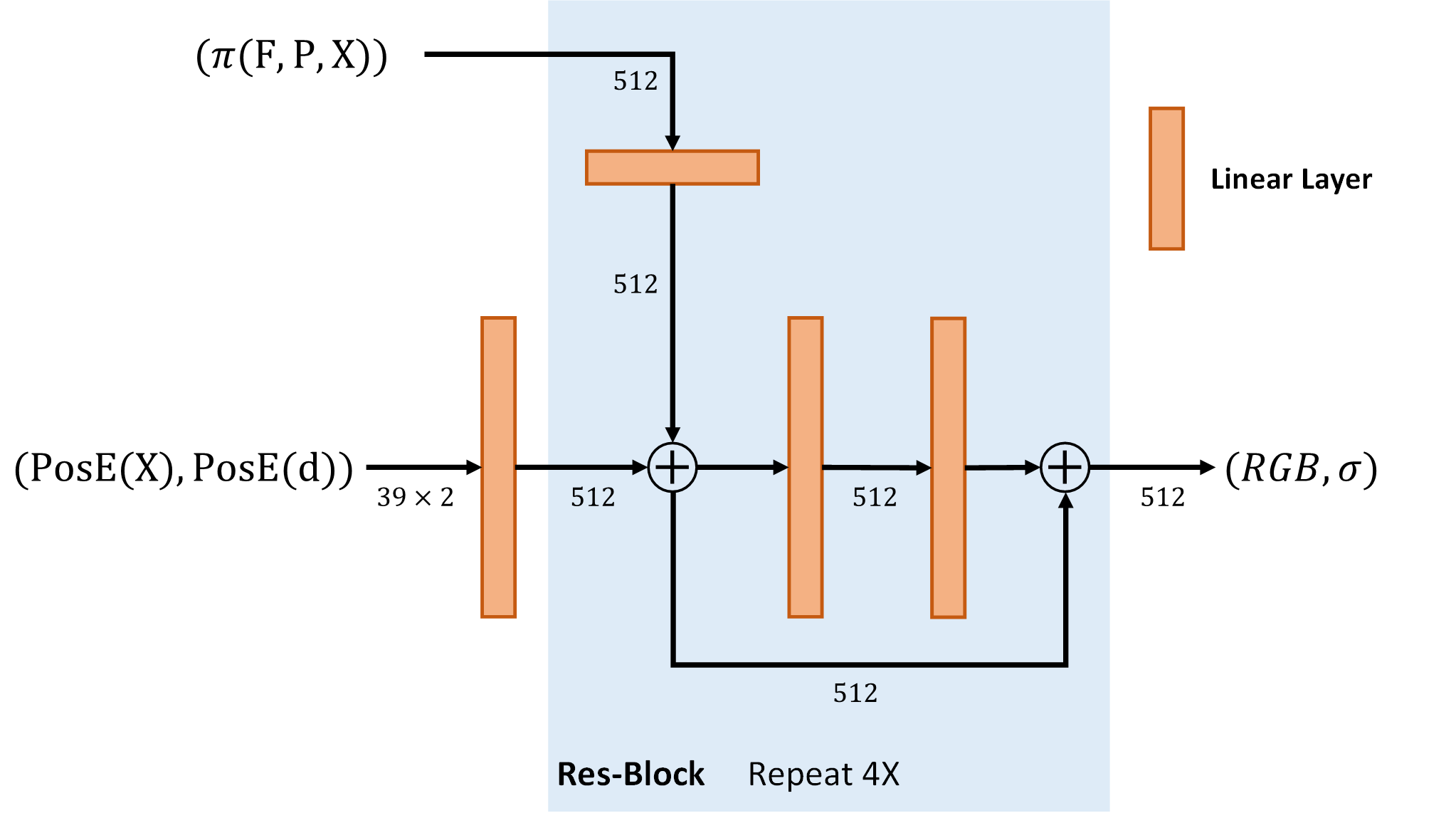}
    \caption{Our MLP structure, composed of four Res-Blocks, takes as input the positionally encoded real-world points (\(\operatorname{PosE}(X), \operatorname{PosE}(d)\)) and the feature-extracted image (\(\pi(\mathrm{F}, \mathrm{P}, \mathrm{X})\)). This design facilitates the integration of spatial and visual data, crucial for generating high-fidelity 3D RGB point clouds.}
    \label{fig:modeldetail}
\end{figure}

In our UNeR3D model, we employed a carefully designed Multi-Layer Perceptron (MLP) architecture, optimized for efficient and accurate 3D reconstruction from 2D imagery. The MLP structure, detailed in Figure \ref{fig:modeldetail}, is built to effectively process positional encodings and image feature vectors, crucial for predicting volumetric density and RGB values of the point cloud.

The entire model was developed using the PyTorch framework and trained on a single Nvidia A800 GPU. This setup was chosen to strike a balance between computational power and efficiency, catering to the resource-intensive nature of our reconstruction tasks. For optimization, we utilized the Adam optimizer, with a learning rate set at \(1 \times 10^{-4}\) and no decay over epochs, ensuring steady progression in learning.

During training, we maintained a batch size of 1 to optimize memory usage and computational performance on the GPU. For each input image, the model generates 256 neural radiance lines, a parameter that balances the need for detailed reconstruction against computational constraints. This configuration allows our model to process a wide range of images, achieving high-precision 3D reconstruction while ensuring efficient use of computational resources.

\begin{figure*}[!ht]
    \centering
    \includegraphics[width=1\linewidth]{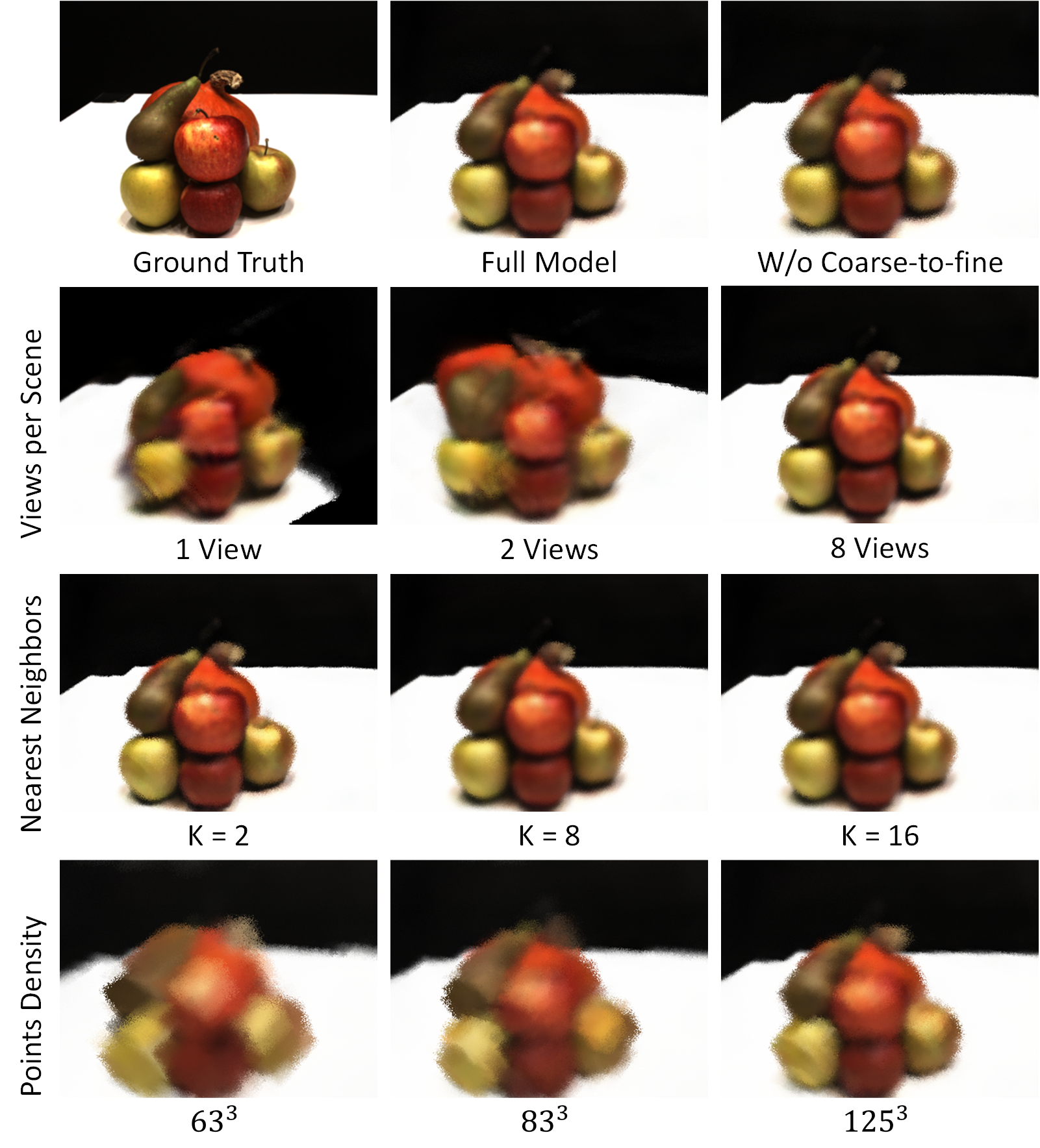}
    \caption{Comparative visualization of the ablation study results. The full model employs a coarse-to-fine structure, processes 4 input views, uses \(k=4\) for knn, and has a resolution of \(250^3\). The results demonstrate that variations in each component of our model configuration impact performance, affirming the efficacy of our approach.}
    \label{fig:abstudy}
\end{figure*}

\section{Dataset Details}
In our experimental setup, we utilized three distinct datasets: ShapeNet\cite{wu20153d}, DTU\cite{aanaes2016large}, and real-world data, each presenting unique characteristics and challenges.

\noindent\textbf{ShapeNet and DTU Datasets:} For both ShapeNet and DTU datasets, camera extrinsics were predetermined to obtain the camera pose matrices, crucial for our model's training and evaluation. In the case of the DTU dataset, inspired by the methodology of NeUS\cite{wang2021neus}, we employed masks to enhance training effectiveness. These masks help isolate the object of interest from complex backgrounds, thereby focusing the model's learning on relevant features.

\noindent\textbf{Real-world Data Collection:} The real-world data, essential for testing our model's generalization capabilities, were sourced from Sketchfab\footnote{\url{https://sketchfab.com/3d-models}}. We specifically selected free-to-use models from this platform, ensuring that the data were used solely for experimental and research purposes, with no commercial intent. The inclusion of these real-world models provides a rigorous testbed for our algorithm, challenging it with data that closely mimics the diversity and complexity encountered in natural environments.

This diverse dataset compilation allowed us to thoroughly evaluate our model across various scenarios, ensuring its robustness and adaptability to both synthetic and real-world conditions.

\section{Future Work Directions}
Our research with UNeR3D has opened multiple avenues for future exploration and development in 3D reconstruction. We envision the following key areas as vital for advancing our work:

\noindent\textbf{Integration with 3D Generative Models:} A promising direction is the amalgamation of our model with 3D generative models, particularly those based on the NeRF framework currently prevalent in text-to-3D research. The implicit nature of NeRF's reconstructions presents challenges in central alignment, an issue that can be potentially mitigated by incorporating our explicit 3D model outputs. This integration could optimize multi-modal alignment processes, enhancing the central alignment and overall fidelity of generated 3D models.

\noindent\textbf{Refinement of 3D Output Quality:} While our model successfully outputs 3D RGB point clouds, there are scenarios where it produces noise, particularly black dots, mainly due to suboptimal separation from the background. Future efforts could focus on refining these outputs, striving for higher quality and resolution. This could involve integrating advanced texture sculpting techniques to elevate the visual quality of the reconstructed models, pushing the boundaries of 3D rendering realism.

\noindent\textbf{Camera Pose Independence:} Our current model still relies on input camera pose matrices. A significant leap would be to unlock the constraints of camera pose, allowing for the input of images from arbitrary angles without needing registration. This development would free the reconstruction process from the limitations of shooting angles, broaden the range of practical applications, and make the technology more accessible and versatile for various real-world scenarios.

These future directions not only aim to enhance the capabilities of our current model but also signify a broader impact on the field of 3D vision, potentially revolutionizing how we approach 3D content creation from 2D data sources.

\end{document}